%% file: vi_for_hmm.tex
\documentclass[10pt,journal,letterpaper,compsoc]{IEEEtran}

\usepackage[utf8]{inputenc}
\usepackage[T1]{fontenc}

\usepackage[noadjust]{cite}      

\usepackage{graphicx}  

\usepackage{amsmath,amssymb}
\usepackage{url}

\usepackage{pifont}

\def\clap#1{\hbox to 0pt{\hss#1\hss}}

\usepackage{subfigure}
\usepackage{booktabs}
\usepackage{multirow}

\usepackage{graphicx} 
\usepackage{amsmath}
\usepackage{dsfont}
\usepackage{tikz}
\usetikzlibrary{fit}

\usepackage{standalone}

\newcommand{\comment}[1]{\marginpar{\color{red}\sloppy\tiny #1}}
\renewcommand{\comment}[1]{}

\usepackage{todonotes}

\newcommand{\bs}[1]{\ensuremath{\boldsymbol{#1}}}
\newcommand{\E}[1]{\ensuremath{\mathds{E}\left[#1\right]}}
\newcommand{\Es}[2]{\ensuremath{\mathds{E}_{#1}\left[#2\right]}}
\newcommand{\mln}{\ensuremath{\mathrm{ln}~}}
\newcommand{\gm}[1]{\ensuremath{\gamma(#1)}}
\newcommand{\avar}[1]{\ensuremath{\upsilon(#1)}}
\newcommand{\bvar}[1]{\ensuremath{\omega(#1)}}

\begin{document}
	%
	\title{Variational Bayesian Inference for Hidden Markov Models With Multivariate Gaussian Output Distributions}
	%
	%
	\author{Christian Gruhl, Bernhard~Sick
		\thanks{C. Gruhl and B. Sick are with the University of Kassel, Department of Electrical Engineering and Computer Science, Wilhelmshoeher Allee 73, 34121 Kassel, Germany (email: {cgruhl,bsick}@uni-kassel.de).}}
	%
	%
	%
	\markboth{Arxiv}{Gruhl, Sick: Variational Bayes for Hidden Markov Models with Multivariate Gaussians.}
	%
	
	
	
	
	\IEEEcompsoctitleabstractindextext{
        \begin{abstract}
            Hidden Markov Models (HMM) have been used for several years in many time series analysis or pattern recognitions tasks. HMM are often trained by means of the Baum-Welch algorithm which can be seen as a special variant of an expectation maximization (EM) algorithm. Second-order training techniques such as Variational Bayesian Inference (VI) for probabilistic models regard the parameters of the probabilistic models as random variables and define distributions over these distribution parameters, hence the name of this technique. VI can also bee regarded as a special case of an EM algorithm. In this article, we bring both together and train HMM with multivariate Gaussian output distributions with VI. The article defines the new training technique for HMM. An evaluation based on some case studies and a comparison to related approaches is part of our ongoing work.
        \end{abstract}
        \begin{keywords}
            Variational Bayesian Inference, Hidden Markov Model, Gaussian-Wishart distribution
        \end{keywords}
    }

	
	\maketitle
	
	\section{Introduction}\label{sec:intro}
    
    \textit{Hidden Markov Models (HMM)} are a standard technique in time series analysis or data mining. Given a (set of) time series sample data, they are typically trained by means of a special variant of an expectation maximization (EM) algorithm, the Baum-Welch algorithm. HMM are used for gesture recognition, machine tool monitoring, or speech recognition, for instance.
    
    \textit{Second-order techniques} are used to find values for parameters of probabilistic models from sample data. The parameters are regarded as random variables, and distributions are defined over these variables. These type of these second-order distributions depends on the type of the underlying probabilistic models. Typically, so called conjugate distributions are chosen, e.g., a Gaussian-Wishart distribution for an underlying Gaussian for which mean and covariance matrix have to be determined. Second-order techniques have some advantages over conventional approaches, e.g.,  
    \begin{itemize}
        \item the uncertainty associated with the determination of the parameters can be numerically expressed and used later,
        \item prior knowledge about parameters can be considered in the parameter estimation process, and
        \item the parameter estimation (i.e., training) process can more easily be controlled (e.g., to avoid singularities),
        \item the training process can easily be extended to automate the search for an appropriate number of model components in a mixture density model (e.g., a Gaussian mixture model).
    \end{itemize}
    If point estimates for parameters are needed, they can be derived from the second-order distributions in a maximum posterior (MAP) approach or by taking the expectation of the second-order distributions. Variational Bayesian Inference (VI), which can also be seen as a special variant of an expectation maximization (EM) algorithm, is a typical second-order approach \cite{Bis06}.        
    
    Although the idea to combine VI and HMM is not completely new and there were already approaches to perform the HMM training in a variational framework (cf. \cite{McGrory09}), typically only models with \textit{univariate} output distributions (i.e.,~scalar values) are considered.
    
    In this article, we bring these two ideas together and propose VI for HMM with \textit{multivariate} Gaussian output distributions. The article defines the algorithm. An in-depth analysis of its properties, an experimental evaluation, and a comparison to related work a are part of our current research.

    \comment{BS@CG: Hier zu dem einen Paper related work, das es gab?}
    
    Section \ref{sec:notationandmodel} introduces the model and the notation we use in our work. Section \ref{sec:ViforHMM} introduces VI for HMM. Finally, Section \ref{sec:conclusion} concludes the article with a summary of the key results and a brief outlook.


\section{Model and Notation}
\label{sec:notationandmodel}

We assume a GMM where each Gaussian is the output distribution of a hidden state.
This is not so simple as it seems on first sight, especially when the Gaussians are overlapping. Thus it is not clear which observation was \textit{generated} by which Gaussian (or by which state).

The GMM can be interpreted as a special instance of a HMM, namely a HMM that consists of a transition matrix with the same transition probabilities from each state to every other state that is similar to the initial state distribution.

The mixing coefficients estimated for the GMM are similar to the starting probabilities of the HMM. \comment{CG@BS: Uff ob das so stimmt? Ich hab es zumindest so implementiert.}

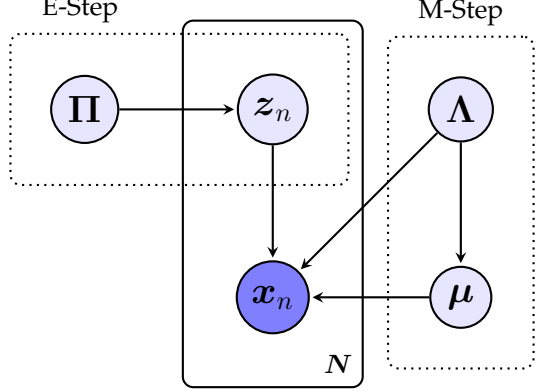
\begin{figure}
	\caption{Graphical Model. Observations $\boldsymbol{x}_n$ are depending on latent variables $\boldsymbol{z}_n$, which are the estimate of the state, as well as the GMM parameters $\boldsymbol{\Lambda}$ (precision matrices) and $\boldsymbol{\mu}$ (mean vector, which also depends on the precision matrix) for the $J$ components. The latent variables $\boldsymbol{z}_n$ have an additional dependency on the transition matrix $\boldsymbol{\Pi}$.\label{fig:graphmod}}
    \centering
        \begin{tikzpicture}[->,>=stealth,shorten >=1pt,auto,node distance=2.5cm, thick,rand node/.style={circle,fill=blue!10,draw,font=\sffamily\Large\bfseries}, main node/.style={circle,fill=blue!50,draw,font=\sffamily\Large\bfseries}]
        
        \node[rand node] (1) {$\boldsymbol{\Pi}$};
        \node[rand node] (2) [right of=1] {$\boldsymbol{z}_n$};
        \node[main node] (3) [below of=2] {$\boldsymbol{x}_n$};
        \node[rand node] (4) [right of=2] {$\boldsymbol{\Lambda}$};
        \node[rand node] (5) [right of=3] {$\boldsymbol{\mu}$};
        
        \node[draw,solid,fit=(2) (3), rounded corners, inner sep=20pt,label={[shift={(-4ex,3.5ex)}]south east:$\boldsymbol{N}$}
        ] {};
        
        \node[draw,dotted,fit=(1) (2), rounded corners, inner sep=15pt,label={[shift={(-8ex,0)}]E-Step}] {};
        \node[draw,dotted,fit=(4) (5), rounded corners, inner sep=15pt,label={M-Step}] {};

        \path[every node/.style={font=\sffamily\small}]
        (1) edge node [left] {} (2)
        (2) edge node [right] {} (3)
        (4) edge node [left] {} (3)
        edge node [left] {} (5)
        (5) edge node [left] {} (3);
        \end{tikzpicture}
\end{figure}

In the remainder of the article, we use the following notation:
\begin{itemize}
    \item $\mathds{E}\left[x\right]$ is the expectation of the random variable $x$,
    \item vectors are denoted with a bold, lowercase symbol e.g., $\boldsymbol{x}$,
    \item matrices are denoted with a bold, uppercase symbol e.g., $\boldsymbol{X}$,
    \item $\boldsymbol{X}$ is the sequence of observations $\boldsymbol{x}_n$, with $1 \leq n \leq N$ and $N = |\boldsymbol{X}|$,
    \item $\boldsymbol{Z}$ is the set of latent variables $\boldsymbol{z}_{n,j}$, with $1 \leq n \leq N$ and $N = |\boldsymbol{X}|$, $1 \leq j \leq J$ and $J$ being the number of states (which is equal to the number of components of the GMM) (Here we use a \textit{1-out-of-K} coding.),
    \item $\boldsymbol{\Theta}$ is the parameter vector/matrix containing all model parameters (including transition probabilities $\boldsymbol{\pi}$, as well as output parameters $\boldsymbol{\mu}, \boldsymbol{\Lambda}$),
    \item $\mathcal{L}$ is the likelihood or its lower bound approximation,
    \item $\boldsymbol{\Pi}$ is the transition matrix with rows $\boldsymbol{\pi}_{i}$,
    \item $\boldsymbol{\pi}_{i}$ are the transition probabilities for state $i$, with $1 \leq i \leq J$ and elements $\pi_{i,j}$,
    \item $\pi_{i,j}$ is the transition probability to move from state $i$ to state $j$, with $1 \leq i,j \leq J$, and
    \item $\pi_j$ is the probability to \textit{start} in state $j$.
\end{itemize}

\comment{BS@CG: Ab hier bis Conclusion nicht bearbeitet (ausser Aenderung der Glederungstiefen).}

\subsection{Hidden Markov Model}

A Hidden Markov Model (HMM)

\comment{CG@CG: Eigentlich wäre es deutlich cooler pro state ein gmm zu haben, dass die Maschkoeffizienten anpasst.}

\section{Variational Inference} \label{sec:ViforHMM}
The direct optimization of p($\boldsymbol{X}|\bs{\Theta}$) is difficult, but the optimization of the complete-data likelihood function p($\boldsymbol{X},\boldsymbol{Z}|\bs{\Theta}$) is significantly easier. We introduce a distribution q($\boldsymbol{Z}$) defined over the latent variables, and we observe that, for any choice of q($\boldsymbol{Z}$), the following decomposition holds
\begin{align}
	\mathrm{ln}~p(\boldsymbol{X}|\bs{\Theta}) &= \mathcal{L}(q,\bs{\Theta})+\mathrm{KL}(q||p)
\end{align}

where we define
\begin{align}
	\mathcal{L}(q,\boldsymbol{\Theta}) &= \int q(\boldsymbol{Z}) \mathrm{ln}\left\lbrace \frac{p(\boldsymbol{X},\boldsymbol{Z})|\boldsymbol{\Theta}}{q(\boldsymbol{Z})}\right\rbrace \mathrm{d}\boldsymbol{Z} \label{eq:lbound}\\
	\mathrm{KL}(q||p) &= - \int q(\boldsymbol{Z}) \mathrm{ln}\left\lbrace \frac{p(\boldsymbol{Z}|\boldsymbol{X},\boldsymbol{\Theta})}{q(\boldsymbol{Z})}\right\rbrace \mathrm{d}\boldsymbol{Z} \label{eq:kl}
\end{align}

The latent variables $\boldsymbol{Z}$ absorbed the model parameters $\boldsymbol{\Theta}$, which are also random variables in this setup. To obtain an optimal model we are interested in maximizing the lower bound with respect to our variational distribution $q$:

\begin{equation}
\underset{q}{\mathrm{argmax}}~\mathcal{L}(q)
\end{equation}

Which is the same as minimizing Eq. \eqref{eq:kl}. Therefore optimum is reached when the variational distribution $q(\boldsymbol{Z})$ matches the conditional (posterior) distribution $p(\boldsymbol{Z}|\boldsymbol{X})$. In that case the $\mathrm{KL(q||p)}$ divergence vanishes since $\mathrm{ln}(1)=0$.

Factorization of the \textit{real} distribution $p$ (see Fig. \ref{fig:graphmod}). N. B.,~only the values of samples $\boldsymbol{x}_n$ are observed.
\begin{align}
p(\boldsymbol{X},\boldsymbol{Z},\boldsymbol{\Pi},\boldsymbol{\mu},\boldsymbol{\Sigma}) &= p(\boldsymbol{X}|\boldsymbol{Z},\boldsymbol{\mu},\boldsymbol{\Lambda})p(\boldsymbol{Z}|\boldsymbol{\Pi})p(\boldsymbol{\Pi})p(\boldsymbol{\mu}|\boldsymbol{\Lambda})p(\boldsymbol{\Lambda})
\end{align}

We assume, that a factorization of the variational distribution $q$ is possible as follows:
\begin{align}
q(\boldsymbol{Z},\boldsymbol{\Pi},\boldsymbol{\mu},\boldsymbol{\Lambda}) &= q(\boldsymbol{Z})q(\boldsymbol{\Pi},\boldsymbol{\mu},\boldsymbol{\Lambda}) \\
&= q(\boldsymbol{Z})q(\boldsymbol{\Pi})q(\boldsymbol{\mu},\boldsymbol{\Lambda}) \\
&= q(\boldsymbol{Z}) \prod_{i=1}^{J} q(\bs{\pi_i}) \prod_{j=1}^{J}q(\boldsymbol{\mu}_j,\boldsymbol{\Lambda}_j)
\end{align}

\subsection{Choice of Distributions}
For each state j, we assign an independent Dirichlet prior distribution for the transition probabilities, so that
\begin{align}
	p(\boldsymbol{\Pi}) &= \prod_{j=1}^{J} Dir(\boldsymbol{\pi}_j|\boldsymbol{\alpha}^{(0)}_{j}) \\
	\boldsymbol{\alpha}^{(0)}_j &= \{\alpha^{(0)}_{j,1},\dots,\alpha^{(0)}_{j,J}\}
	\end{align}
	

The variational posterior distributions for the model parameters turn out to have the following form: \comment{CG@CG wirklich posteriors?}

\begin{align}
q(\boldsymbol{\Pi}) &= \prod_{j=1}^{J} Dir(\boldsymbol{\pi}_j|\boldsymbol{\alpha}_{j}) \\
\boldsymbol{\alpha}_j &= \{\alpha_{j,1},\dots,\alpha_{j,J}\} 
\end{align}


The means are assigned independent univariate Gaussian conjugate prior distributions, conditional on the precisions. The precisions themselves are assigned independent Wishart prior distributions:
\begin{align}
	p(\bs{\mu},\bs{\Lambda}) &= p(\bs{\mu}|\bs{\Lambda})p(\bs{\Lambda}) \\
	&= \prod_{j=1}^{J}\mathcal{N}(\bs{\mu}_j|\bs{m}_0,(\beta_0\bs{\Lambda}_j)^{-1}) \cdot \mathcal{W}(\bs{\Lambda}_j|\bs{W}_0,\nu_0) \label{eq:p_mu_Lambda}
\end{align}

The variational posterior distributions for the model parameters are as fallows (application of Bayes theorem):

\begin{align}
q(\bs{\mu}_j,\bs{\Lambda}_j) &= \mathcal{N}(\bs{\mu}_j|\bs{m}_j,(\beta_j\bs{\Lambda}_j)^{-1}) \cdot \mathcal{W}(\bs{\Lambda}_j|\bs{W}_j,\nu_j)
\end{align}

The variational posterior for q(Z) will have the form:
\begin{align}
q(\bs{Z}) &\propto \prod_{n=1}^{M} \prod_{j=1}^{J} ({b_{n,j}})^{z_{n,j}} \prod_{n=1}^{N} \prod_{j=1}^{J} \prod_{s=1}^{J} (a_{j,s})^{z_{n,j}\cdot z_{(n+1),s}} \label{eq:q_Z}
\end{align}

Which is identical to the one given by McGrory et al. in \cite{McGrory09}. The expected logarithm for Eq. \eqref{eq:q_Z} can be derived to:

\begin{align}
\E{\mln q(\bs{Z})} &= \sum_{n=1}^{N} \sum_{j=1}^{J} \gm{z_{n,j}} \E{\mln p(\bs{x}_n|\bs{\mu}_j,\bs{\Lambda}^{-1}_j)} \notag \\
                   &+ \sum_{n=1}^{N-1} \sum_{j=1}^{J} \sum_{s=1}^{J} \xi(z_{n,j},z_{n+1,s}) \E{\mln \tilde{\pi}_{j,s}}
\end{align}

The distribution over $\bs{Z}$ given the transition table $\bs{\Pi}$ expands to: \comment{CG: Dies ist eine von 'unseren' Leistungen.}

\begin{align}
p(\bs{Z}|\bs{\Pi}) &= p(\bs{z}_1|\bs{\pi}) \prod_{n=2}^{N} (p(\bs{z}_n|\bs{z}_{n-1}))^{z_{n-1,j}.z_{n,s}} \\
&= \bs{\pi} \prod_{n=2}^{N}(\bs{a}_{n-1,n})^{z_{n-1,j}.z_{n,s}} \label{eq:lh_P_Z_Pi}
\end{align}

and its expected value to:

\begin{align}
\E{\mln p(\bs{Z}|\bs{\Pi})} &= \sum_{j=1}^{J} \pi_j \notag \\ 
&+ \sum_{n=2}^{N} \sum_{j=1}^{J} \sum_{s=1}^{J} \xi(z_{n-1,j},z_{n,s}) \cdot \E{\mln \tilde{\pi}_{j,s}}
\end{align}

For Eq. \eqref{eq:lh_P_Z_Pi} confer Eq. \eqref{eq:a_j_s} and Eq. \eqref{eq:pi_j}.\\

\subsection{E-Step}

The expectation step (E-Step) uses the Baum Welch algorithm to estimate the latent variables for all observations in the sequence.

The latent variables $\gm{\bs{z}_{n,j}}$ denotes the probability that the observation at time step $n$ was generated by the $j$-th component of the model.
\begin{align}
\gm{\bs{z}_{n}} = \E{\bs{z}_{n}} &= p(\bs{z}_n|\bs{X}) = \frac{\avar{\bs{z}_n}\bvar{\bs{z}_n}}{\sum_{\bs{z} \in \bs{Z}}\avar{\bs{z}}\bvar{\bs{z}}} \\
\gm{z_{n,j}} = \E{z_{n,j}} &= \frac{\avar{z_{n,j}}\bvar{z_{n,j}}}{\sum_{k=1}^{J}\avar{z_{n,k}}\bvar{z_{n,k}}}
\end{align}

The transition probabilities  $\xi(z_{n-1,j},z_{n,s})$ express the uncertainty how likely it is, that a transition from state $j$ to $s$ has happened if observation $\bs{x}_{n-1}$ was generated by the $j$-th component and the $n$-th observation by the $s$-th component.
\begin{align}
\xi(\bs{z}_{n-1},\bs{z}_n) &\propto \avar{\bs{z}_{n-1}}p(\bs{z}_n|\bs{z}_{n-1})p(\bs{x}_n|\bs{z}_n)\bvar{\bs{z}_{n}} \\
\xi(z_{n-1,j},z_{n,s}) &\propto \avar{z_{n-1,j}}a_{j,s} b_{n,s}\bvar{z_{n,s}} \\
\xi(z_{n-1,j},z_{n,s}) &= \frac{\avar{z_{n-1,j}}a_{j,s} b_{n,s}\bvar{z_{n,s}}}{\sum_{k=1}^{J}\sum_{l=1}^{J}\avar{z_{n-1,k}}a_{k,l} b_{n,l}\bvar{z_{n,l}}}
\end{align}

with

\begin{align}
a_{j,s} &= \mathrm{exp}\left\{ \E{\mln \tilde{\pi}_{j,s}}\right\} \label{eq:a_j_s} \\
b_{n,j} &= \mathrm{exp}\left\{ \E{\mln p(\bs{x}_n|\bs{\mu}_j,\bs{\Lambda}_j)}\right\}
\end{align}	

and the popular Baum Welch algorithm:

\begin{align}
\avar{z_{1,j}} &= \pi_j b_{1,j} \\
\avar{z_{n,j}} &= b_{n,j} \sum_{k=1}^{J}\avar{z_{n-1,k}} \\
\bvar{z_{N,j}} &= 1\\
\bvar{z_{n,j}} &= \sum_{s=1}^{J}\bvar{z_{n+1,s}} \cdot a_{j,s} \cdot b_{n+1,s}
\end{align}

Note that we substituted the common function names to $\alpha \rightarrow \nu$ and $\beta \rightarrow \omega$, since those symbols are already occupied by other definitions.

\begin{align}
\E{\mln p(\bs{x}_n|\bs{\mu}_j,\bs{\Lambda}^{-1}_j)} = \frac{1}{2} \Es{\bs{\Lambda}}{\mln |\bs{\Lambda}_j|} - \frac{\mln(2\pi)D}{2} \notag \\
- \frac{1}{2} \Es{\bs{\mu},\bs{\Lambda}}{(\bs{x}_n-\bs{\mu}_j)^T\bs{\Lambda}_j(\bs{x}_n-\bs{\mu}_j)}
\end{align} \comment{CG: Somewhat related to Eq. \eqref{eq:p_mu_Lambda}}

\begin{align}
\Es{\bs{\Lambda}_j}{\mln |\bs{\Lambda}_j|} &= \sum_{d=1}^{D} \psi(\frac{\nu_j + 1 - d}{2}) + D \mln 2 + \mln |\bs{W}_j| \label{eq:E_log_det_Lambda}\\
&\Es{\bs{\mu}_j,\bs{\Lambda}_j}{(\bs{x}_n-\bs{\mu}_j)^T\bs{\Lambda}_j(\bs{x}_n-\bs{\mu}_j)} \notag \\
&= D\beta^{-1}_j + \nu_j(\bs{x}_n-\bs{m}_j)^T\bs{W}_j(\bs{x}_n-\bs{m}_j) \label{eq:E_dist}
\end{align}
Eq. \eqref{eq:E_log_det_Lambda} and Eq. \eqref{eq:E_dist} are taken from \cite{Bis06}.

Where $D = |\bs{x}|$ is the dimensionality of the observed data.

The conditional probability that a given observation $\bs{x}_n$ is \textit{generated} by the $j$\textit{th} state (that means the probability of $z_{n,j} = 1$ ) is given by:
\begin{equation}
\gm{z_{n,j}} = \E{\bs{z}_{n,j}} = \sum_{\bs{z}} \gm{\bs{z}}\cdot z_{n,j}
\end{equation}
Estimating the initial probabilities $\bs{\pi}$ that the model starts in state $j$.
\begin{align}
\pi_j &= \frac{\gm{z_{1,j}}}{\sum_{k=1}^{J}\gm{z_{1,k}}} = \gm{z_{1,j}=1} \label{eq:pi_j}
\end{align}

At the beginning of the sequence there is no predecessor state, thus we can directly use the estimated latent variable as prior probability for the $j$-th state.

\subsection{M-Step} 

The Maximization step:

\begin{equation}
N_j = \sum_{n=1}^{N} \gm{z_{n,j}}
\end{equation}

\begin{equation}
\bar{\bs{x}}_j = \frac{1}{N_j}\sum_{n=1}^{N}\gm{z_{n,j}}\bs{x}_n
\end{equation}

\begin{equation}
\bs{S}_j = \frac{1}{N_j}\sum_{n=1}^{N}\gm{z_{n,j}}(\bs{x}_n-\bar{\bs{x}}_j)(\bs{x}_n-\bar{\bs{x}}_j)^T
\end{equation}

\begin{align}
\beta_{j} &= \beta_{0} + N_j \\
\nu_{j} &= \nu_{0} + N_j \\
\bs{m}_{j} &= \frac{1}{\beta_{j}}(\beta_{0}\bs{m}_{0}+\bs{\bar{x}}_{j}N_j)
\end{align}

\begin{align}
\bs{W}_{j}^{-1} &= \bs{W}_{0}^{-1} + N_j\bs{S}_{j} \notag \\
&+\frac{\beta_{0}N_j}{\beta_{0}+N_j} (\bs{\bar{x}}_{j}-\bs{m}_{0})(\bs{\bar{x}}_{j}-\bs{m}_{0})^{T}
\end{align}

All equations are based on the Variational Mixture of Gaussians \cite{Bis06} and adjusted for our hidden Markov model.

Maximizing the hyper distribution for the transition matrix:
\begin{equation}
\alpha_{j,s} = \alpha^{(0)}_{j,s} + \sum_{n=1}^{N-1} \xi(z_{n,j},z_{(n+1),s})
\end{equation}
for $ 1 \leq j,s \leq J $ and $\boldsymbol{\alpha}^{(0)}_j$ being the prior values for state \textit{j}. 

The hyper parameters for the initial starting probabilities are maximized as fallows:
\comment{CG@BS das sind die Mischkoeffizienten im normalen GMM.}
\begin{equation}
\alpha_j = \alpha_0 + N_j
\end{equation}

\subsection{Variational LowerBound}

To decide whether the model has converged we consult the change of the likelihood function of the model. If the model does not change anymore (or only in very small steps) the likelihood of multiple successive training iterations will be nearly identical. 
The calculation of the actual likelihood is too hard (is it even possible?) to be practicable, to cumbersome, this we use a lower bound approximation for the likelihood.

For the variational mixture of Gaussians, the lower bound is given by

{\scriptsize
    \begin{equation}
    \mathcal{L} = \sum_{\bs{Z}} \int\int\int q(\bs{Z},\bs{\Pi},\bs{\mu},\bs{\Lambda}) \mln\left\{\frac{p(\bs{X},\bs{Z},\bs{\Pi},\bs{\mu},\bs{\Lambda})}{q(\bs{Z},\bs{\Pi},\bs{\mu},\bs{\Lambda})}\right\} \mathrm{d}\bs{\Pi}\mathrm{d}\bs{\mu}\mathrm{d}\bs{\Lambda} 
    \end{equation}
}

Wich is in our case:

\begin{align}
\mathcal{L} &= \E{\mln p(\bs{X},\bs{Z},\bs{\Pi},\bs{\mu},\bs{\Lambda})} - \E{\mln  q(\bs{Z},\bs{\Pi},\bs{\mu},\bs{\Lambda})} \\
&=    \E{\mln p(\bs{X}|\bs{Z},\bs{\mu},\bs{\Lambda})}
+ \E{\mln p(\bs{Z}|\bs{\Pi})}  \notag \\
&~~~+ \E{\mln p(\bs{\Pi})}s
+ \E{\mln p(\bs{\mu},\bs{\Lambda})} 
- \E{\mln q(\bs{Z})} \notag \\
&~~~- \E{\mln q(\bs{\Pi})}
- \E{\mln q(\bs{\mu},\bs{\Lambda})}
\end{align}

The lower bound $\mathcal{L}$ is used to detect convergence of the model, i.e., approaching of the best (real) parameters of the underlying distribution.

\begin{align}
&\E{\mln p(\bs{\Pi}))} \notag \\
&= \sum_{j=1}^{J} \left\{\mln (C(\bs{\alpha}^{(0)}_j))+ \sum_{s=1}^{J}(\alpha^{(0)}_{j,s}-1)\cdot\E{\mln \tilde{\pi}_{j,s}}\right\} \\
&\E{\mln q(\bs{\Pi}))} \notag \\
&= \sum_{j=1}^{J} \left\{\mln (C(\bs{\alpha}_j))+ \sum_{s=1}^{J}(\alpha_{j,s}-1)\cdot\E{\mln \tilde{\pi}_{j,s}}\right\} \\
&\E{\mln \tilde{\pi}_{j,s}} = \psi(\alpha_{j,s}) - \sum_{k=1}^{J}\psi(\alpha_{j,k}) \label{eq:ElnPi} \\ 
&\E{\mln p(\bs{X}|\bs{Z},\bs{\mu},\bs{\Lambda})} = \frac{1}{2} \sum_{j=1}^{J} N_j \{ \mln \E{\mln |\bs{\Lambda}_j|}  \notag \\
&- D\beta^{-1}_j - \nu_j \mathrm{Tr}(\bs{S}_j\bs{W}_j) - \nu_j(\bs{\bar{x}}_{j}-\bs{m}_{j})^T\bs{W}_j(\bs{\bar{x}}_{j}-\bs{m}_{j}) \notag \\
&- D \mln(2\pi)\} \\
&\E{\mln p(\bs{\mu},\bs{\Lambda})} = \frac{1}{2} \sum_{j=1}^{J} \{ D \mln (\beta_0/2\pi) + \E{\mln |\bs{\Lambda}_j|} \notag \\
&- \frac{D\beta_0}{\beta_j} - \beta_0\nu_j(\bs{m}_{j}-\bs{m}_{0})^T\bs{W}_j(\bs{m}_{j}-\bs{m}_{0}) \notag \\
&+ J~\mln \mathrm{B}(\bs{W}_0,\nu_0) + \frac{\nu_0 - D - 1}{2}\sum_{J}^{j=1}\E{\mln |\bs{\Lambda}_j|} \notag \\
&-\frac{1}{2}\sum_{j=1}^{J}\nu_j \mathrm{Tr}(\bs{W}^{-1}_0\bs{W}_j) \}\\
&\E{\mln q(\bs{\mu},\bs{\Lambda})} = \sum_{j=1}^{J} \{ \frac{1}{2}\E{\mln |\bs{\Lambda}_j|} + \frac{D}{2} \mln (\beta_j/2\pi) \notag \\
&-\frac{D}{2} - \mathrm{H}\left[q(\bs{\Lambda}_j)\right] \} \\
&\mathrm{H}\left[q(\bs{\Lambda}_j)\right] = \mln \mathrm{B}(\bs{W}_j,\nu_j) - \frac{\nu_j - D - 1}{2}\E{\mln |\bs{\Lambda}_j|} \notag \\
&+ \frac{\nu_jD}{2}
\end{align}

%
%


%

\subsection{Choice of Hyper Parameters}

We can either choose the priors $\alpha_{j,s}^0$ for the transitions on random (with a seed) or make some assumptions and use that for initialization e.g.: \comment{Das ist nach dem Entfernen von Komponenten ggf. problematisch, da dann das Verbleiben im Zustand verstaerkt wird.}

\begin{align}
	\alpha_{j,s}^0 = \begin{cases}
        0.5,& \text{if } j = s\\
    	\frac{1}{2J},              & \text{otherwise}
	\end{cases}
\end{align}

Which means that the probability to stay in a state is always $.5$ and transitions to the other states is equally likely. \comment{Ich wuerde so einen Ansatz bevorzugen, da wir so nicht noch mehr Parameter mit einbringen.}

The prior for the starting states $\alpha_j^0 = \alpha^0$ is the same for all states 
.

Do we need to include actual prior knowledge when using VI approaches?
No, the main advantage in relying on a training that is based on variational Bayesian inference is, that the introduction of the distributions over the model parameters prevents us from running into local minima. Especially those that arise when a component (or a state) collapses over a single observation (or multiple observations with identical characteristics). In that case the variance approaches $0$ (and the mean $\infty$ since $\int p(x) \mathrm{d}x = 1$) and would normally dramatically increase the likelihood for the model -- this is also known as \textit{singularity} (and one of the known drawbacks of \textit{normal} EM).
Using the 2nd order approaches prevents this by a low density in the parameter space where the variance approaches $0$ (Therefore $p(x|\sigma)\rightarrow \infty$ is attenuated by a low density for $p(\sigma)$). 

%

\section{Conclusion and Outlook}\label{sec:conclusion}

In this article, we adapted the concept of second-order training techniques to Hidden Markov Models. A training algorithm has been defined following the ideas of Variational Bayesian Inference and the Baum-Welch algorithm. 

Part of our ongoing research is the evaluation of the new training algorithm using various benchmark data, the analysis of the computational complexity of the algorithm as well as the actual run-times on these benchmark data, and a comparison to a standard Baum-Welch approach.

In our future work we will extend the approach further by allowing different discrete and continuous distributions in different dimensions of an output distribution. We will also used the HMM trained with the new algorithm for anomaly detection (in particular for the detection of temporal anomalies). For that purpose, we will extend the technique we have proposed for GMM in  \cite{GS16}.


\section*{Acknowledgment}

This work was supported by the German Research Foundation (DFG) under grant number SI 674/9-1 (project CYPHOC).

\input{appendix.tex}

\comment{BS@CG: Was ist mit diesen Referenzen? Mindestens noch 1--2 fuer HMM ergaenzen, z.B. das Rabiner-Tutorial. Und geeignet einbauen.}

\bibliographystyle{IEEEtran}
\bibliography{vi_for_hmm}

\begin{IEEEbiography}[{\includegraphics[width=1in,height=1.25in,clip,keepaspectratio]{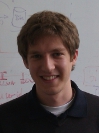}}]{Christian Gruhl}
	received his M.Sc. in Computer Science in 2014 at the University of Kassel with Honors. Currently he is working towards his Ph.D. as a research assistant at the Intelligent Embedded Systems lab of Prof. Sick. His research interests focuses on 2nd order training algorithms for machine learning and its applications in cyber physical systems.
	
	
\end{IEEEbiography}

\begin{IEEEbiography}[{\includegraphics[width=1in,height=1.25in,clip,keepaspectratio]{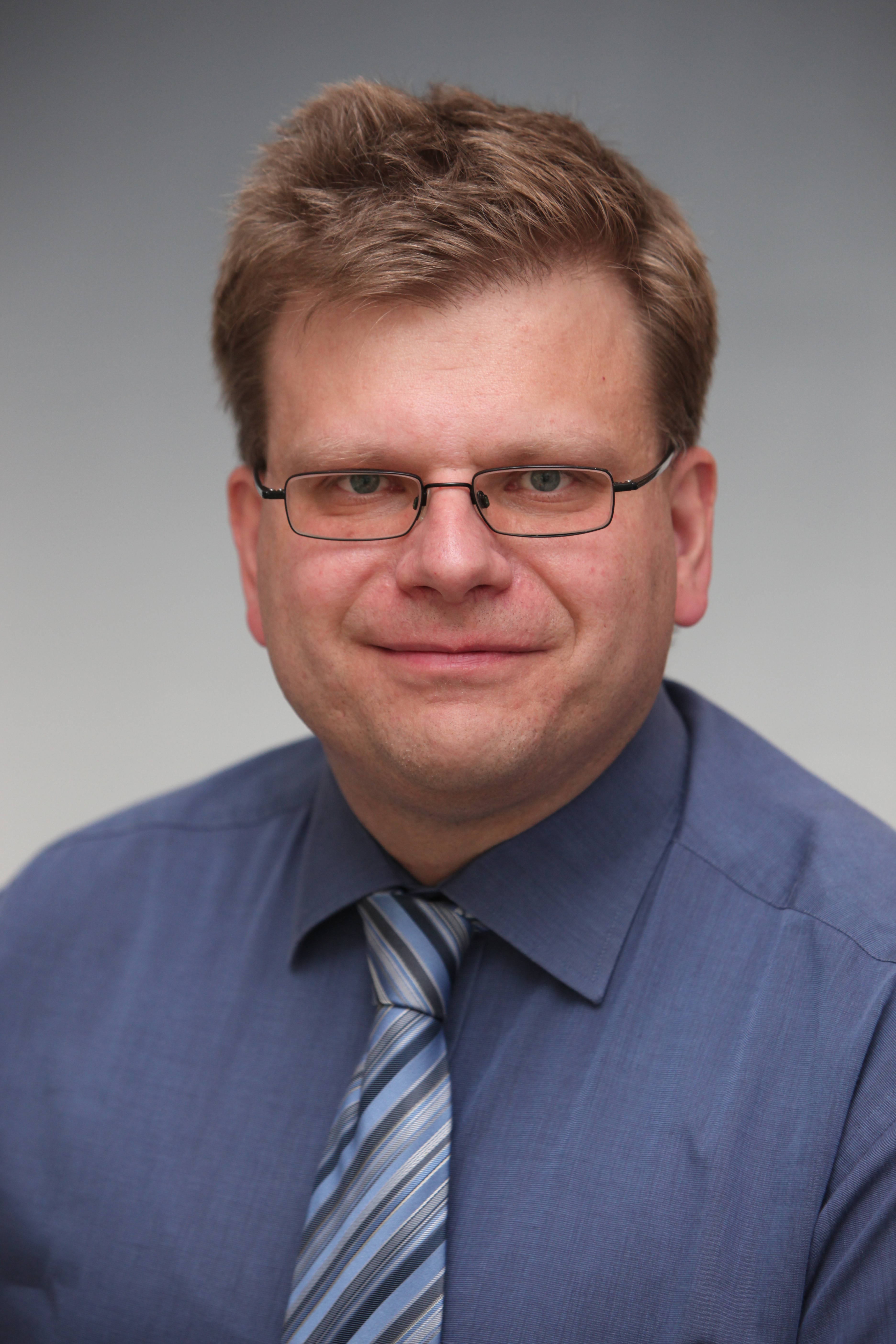}}]{Bernhard Sick}
	received a diploma (1992, M.Sc.\ equivalent), a Ph.D. degree (1999), and a
	``Habilitation'' degree (2004), all in computer science, from the University
	of Passau, Germany. Currently, he is full professor for Intelligent Embedded
	Systems at the Faculty for Electrical Engineering and Computer
	Science of the University of Kassel, Germany. There, he is conducting
	research in the areas Autonomic and Organic Computing and Technical Data Analytics with applications in biometrics, intrusion detection, energy management, automotive engineering, and others. Bernhard Sick
	authored more than 100 peer-reviewed publications in these areas. He is
	a member of IEEE (Systems, Man, and Cybernetics Society, Computer
	Society, and Computational Intelligence Society) and GI (Gesellschaft
	fuer Informatik). Bernhard Sick is associate editor of the IEEE Transactions
	on Cybernetics; he holds one patent
	and received several thesis, best paper, teaching, and inventor awards.
	
\end{IEEEbiography}

%


\vfill


\end{document}

%% file: appendix.tex
\section*{Appendix}

\subsection*{Dirichlet}
\begin{equation}
Dir(\bs{\mu}|\bs{\alpha}) = C(\bs{\alpha}) \prod_{k=1}^{K} \mu_k^{\alpha_{k-1}}
\end{equation}
with constraints
\begin{align}
\sum_{k=1}^{K} \mu_j &= 1 \\
0 \leq \mu_k &\leq 1 \\
|\bs{\mu}| = |\bs{\alpha}| &= K
\end{align}

\begin{align}
\hat{\alpha} &= \sum_{k=1}^K \alpha_k \\
C(\bs{\alpha}) &= \frac{\Gamma(\hat{\alpha})}{\prod_{k=1}^{K}\Gamma(\alpha_k)} \\
\mathrm{H}[\mu] &= -\sum_{k=1}^K(\alpha_k-1)\{\psi(\alpha_k)-\psi(\hat{\alpha})\}- \mln \mathrm{C}(\bs{\alpha})
\end{align}
Digamma function:
\begin{align}
\Psi(a) \equiv \frac{\mathrm{d}}{\mathrm{d}a}\mln\Gamma(a)
\end{align}
Gamma function:
\begin{align}
\Gamma(x) \equiv \int_0^{\infty}u^{x-1}e^{-u}\mathrm{d}u
\end{align}

\subsection*{Normal and Wishart}

\begin{align}
\mathcal{N}(\bs{x}|\bs{\mu},\bs{\Lambda}) &= \frac{1}{(2\pi)^{D/2}} \frac{1}{|\bs{\Lambda}^{-1}|^{1/2}} \notag \\
&\cdot \mathrm{exp}\left\{\frac{1}{2} (\bs{x}-\bs{\mu})^T\bs{\Lambda}(\bs{x}-\bs{\mu})\right\} \\
\mathcal{W}(\bs{\Lambda}|\bs{W},\nu) &= \mathrm{B}(\bs{W},\nu)|\bs{\Lambda}|^\frac{(\nu-D-1)}{2} \notag \\ 
&\cdot \mathrm{exp}\left\{\frac{1}{2}\mathrm{Tr}(\bs{W}^{-1}\bs{\Lambda})\right\}
\end{align}
where
\begin{align}
\mathrm{B}(\bs{W},\nu) &= |\bs{W}|^{-\nu/2}  \notag \\
&\cdot \left(2^{\nu D/2}\pi^{D(D-1)/4}\prod_{i=1}^D\Gamma\left(\frac{\nu+1-i}{2}\right)\right)^{-1}
\end{align}

\subsection*{Gaussian-Wishart}

\begin{align}
p(\bs{\mu}, \bs{\Lambda}|\bs{\mu_{0}},\beta, \bs{W}, \nu) = \mathcal{N}(\bs{\mu}|\bs{\mu_{0}},(\beta\bs{\Lambda})^-1) \mathcal{W}(\bs{\Lambda}|\bs{W}, \nu)
\end{align}

This is the conjugate prior distribution for a multivariate Gaussian $\mathcal{N}(\bs{x}|\bs{\mu,\Lambda})$ in which both the mean $\mu$ and precision matrix $\Lambda$ are unknown, and is also called the Normal-Wishart distribution.